\documentclass{article}

\PassOptionsToPackage{numbers, compress}{natbib}

\usepackage[preprint]{neurips_2023}

\usepackage[utf8]{inputenc} %
\usepackage[T1]{fontenc}    %
\usepackage{url}            %
\usepackage{amsfonts}       %
\usepackage{nicefrac}       %
\usepackage{microtype}      %

\usepackage{pifont}

\usepackage{booktabs}
\usepackage{multirow}
\usepackage{graphicx}
\usepackage{float}
\usepackage{mathtools}
\usepackage{ifthen}
\usepackage{bm}
\usepackage{fp}
\usepackage{siunitx}
\usepackage{amsthm}
\usepackage{caption}

\usepackage{subcaption}
\usepackage{xspace}
\usepackage{xcolor}
\usepackage{amsmath, amsthm, amssymb}
\usepackage{pifont}

\usepackage{wrapfig}
\usepackage{pifont}

\usepackage{algorithm}
\usepackage{algorithmicx}
\usepackage[noend]{algpseudocode}
\usepackage{comment}
\algnewcommand\algorithmicinput{\textbf{Input:}}
\algnewcommand\Input{\item[\algorithmicinput]}
\algnewcommand\algorithmicoutput{\textbf{Output:}}
\algnewcommand\Output{\item[\algorithmicoutput]}
\algnewcommand{\LineComment}[1]{\State \(\triangleright\) #1}

\usepackage[hidelinks,breaklinks,colorlinks]{hyperref}
\hypersetup{
  linkcolor={blue!70!black},
  citecolor={red!70!black},
  urlcolor={blue!70!black}
}

\DeclareMathOperator*{\argmin}{arg\,min}

\usepackage{hyperref}

\theoremstyle{plain}
\newtheorem{theorem}{Theorem}[section]

\theoremstyle{definition}
\newtheorem{definition}[theorem]{Definition}

\theoremstyle{remark}

\newcommand{\sys}{\mbox{\textsc{RONAN}}\xspace}

\def\Snospace~{\S{}}

\usepackage{color}

\usepackage{soul}
\usepackage{xcolor}
\definecolor{DarkGreen}{RGB}{34,139,34}

\usepackage[super]{nth}

\title{Alteration-free and Model-agnostic Origin Attribution of Generated Images}

\author{%
  Zhenting Wang\thanks{Work
partially done during Zhenting Wang’s internship at Sony AI.} \\
  Rutgers University \\
  \texttt{zhenting.wang@rutgers.edu}
  \And
  Chen Chen \\
  Sony AI \\
  \texttt{ChenA.Chen@sony.com} \\
  \And
  Yi Zeng \\
  Virginia Tech \\
  \texttt{yizeng@vt.edu} \\
  \And
  Lingjuan Lyu\thanks{Corresponding Author} \\
  Sony AI \\
  \texttt{Lingjuan.Lv@sony.com} \\
  \And
  Shiqing Ma \\
  Rutgers University \\
  \texttt{sm2283@rutgers.edu} \\
}

\begin{document}

\maketitle

\begin{abstract}

Recently, there has been a growing attention in image generation models. However, concerns have emerged regarding potential misuse and intellectual property (IP) infringement associated with these models. Therefore, it is necessary to analyze the origin of images by inferring if a specific image was generated by a particular model, i.e., origin attribution.
Existing methods are limited in their applicability to specific types of generative models and require additional steps during training or generation. This restricts their use with pre-trained models that lack these specific operations and may compromise the quality of image generation. 
To overcome this problem, we first develop an alteration-free and model-agnostic origin attribution method via input reverse-engineering on image generation models, i.e., inverting the input of a particular model for a specific image. 
Given a particular model, we first analyze the differences in the hardness of reverse-engineering tasks for the generated images of the given model and other images. Based on our analysis, we propose a method that utilizes the reconstruction loss of reverse-engineering to infer the origin. Our proposed method effectively distinguishes between generated images from a specific generative model and other images, including those generated by different models and real images.

\end{abstract}
\section{Introduction}\label{sec:intro}

In recent years, there has been a rapid evolution in image generation techniques.
With the advances in visual generative models,
images can now be easily created with high quality and diversity~\cite{radford2015unsupervised,song2020denoising,rombach2022high,song2023consistency}. There are three important milestones in the field of image generation and manipulation, i.e., Generative Adversarial Networks (GAN)~\cite{Goodfellow2014GenerativeAN}, Variational AutoEncoders (VAE)~\cite{kingma2013auto}, and diffusion models~\cite{ho2020denoising}.
Various image generation models are built based on these three models~\cite{karras2019style,karras2020analyzing,karras2020training,van2017neural,zhu2020domain,dalle2} to make the AI-generated images more realistic.\begin{wrapfigure}{R}{0.58\textwidth}
		\centering
		\footnotesize
		\includegraphics[width=.5\columnwidth]{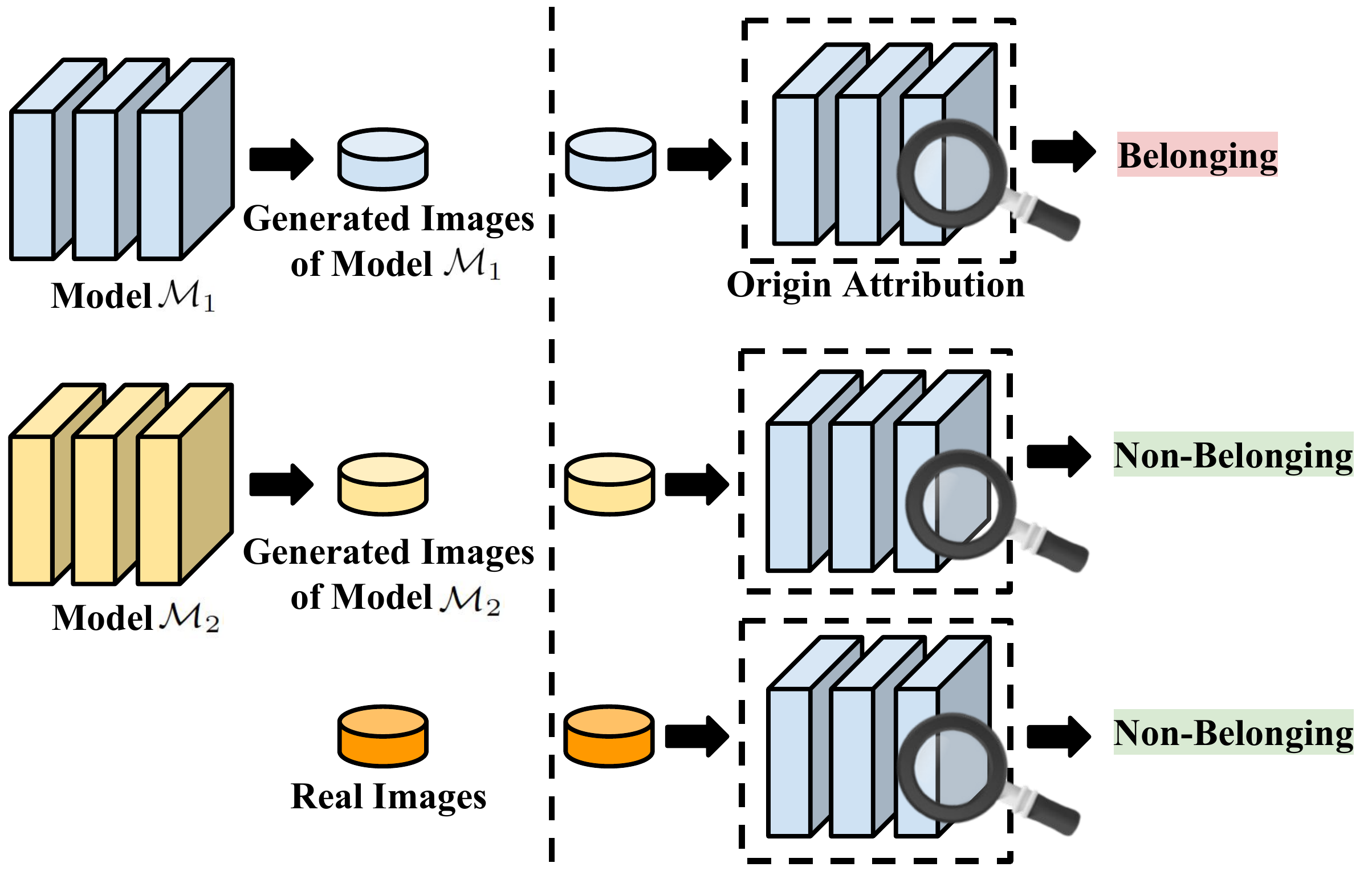}
	\caption{Illustration for origin attribution problem. The origin attribution algorithm aims to judging whether the given images belong to a particular model, i.e., Model \(\mathcal{M}_1\).
    }\label{fig:intro}
    \vspace{-0.3cm}
\end{wrapfigure}

With its wide adoption, the
security and privacy of machine learning models becomes critical~\cite{hayes2017logan,carlini2023extracting,zhao2021multi,tao2022backdoor,chen2020gan,wang2022rethinking,ong2021protecting,wang2022bppattack}.
One severe and important issue is the potential misuse and intellectual property (IP) infringement of image generation models~\cite{zhao2021multi,ong2021protecting}. Users may generate malicious images containing inappropriate or biased content using these models and distribute them online. Furthermore, trained models may be used without authorization, violating the model owner's intellectual property. For example, malicious users may steal the model's parameters file and use it for commercial purposes. 
Others may create AI-generated images and falsely claim them as their own artwork (e.g., photos and paintings) to gain recognition, which also violates the model's IP. 
Therefore, it is essential to track the origin of AI-generated images. The origin attribution problem is to identify whether a specific image is generated by a particular model. As shown in \autoref{fig:intro}, assuming we have a model \(\mathcal{M}_1\) and its generated images, the origin attribution algorithm's objective is to flag an image as belonging to model \(\mathcal{M}_1\) if it was generated by that model. On the other hand, the algorithm should consider the image as non-belonging if it was created by other models (e.g., \(\mathcal{M}_2\) in \autoref{fig:intro}) or if it is a real image.

One existing way to infer the source of specific images is image watermarking.
It works by embedding ownership
information in carrier images to verify the owner's identity
and
authenticity~\cite{swanson1996transparent,luo2009reversible,pereira2000robust,tancik2020stegastamp}.
The image watermarking-based method requires an additional modification to the generation results in a post-hoc manner, which may impair the generation quality.
Also, it might not necessarily reflect the use of a particular model in the process of generating an image, which can reduce its credibility as evidence in lawsuits. Furthermore, the watermark can also be stolen, and malicious users can engage in criminal activities and disguise their identities using the stolen watermark. 
Another approach to identifying the
source models of the generated
samples~\cite{yu2019attributing,ding2021does,yu2021artificial,yu2022responsible} is injecting fingerprinting into the models (e.g., modifying the model architecture) and training a supervised classifier to detect the fingerprints presented in the image. 
While their goal is similar to ours, these methods have several limitations. Firstly, they require extra operations during the model training phase, and they cannot be applied on pre-trained models without additional operations, such as modifying the model architecture~\cite{yu2021artificial,yu2022responsible}. Secondly, since these methods modify the training or inference process of generative models, the model's generation performance may be affected. Thirdly, these previous studies mainly focus on a particular kind of generative model, i.e., GAN~\cite{Goodfellow2014GenerativeAN}. In contrast, our goal is to develop an origin attribution approach for different types of generative models, including diffusion models~\cite{song2023consistency,rombach2022high} (model-agnostic), without requiring any extra operations in the training phase and image generation phase (alteration-free). We summarize the differences between our method and existing methods in \autoref{tab:intro_diff}.

\begin{wraptable}{r}{0.55\linewidth}
\vspace{-0.3cm}
\centering
\scriptsize
\setlength\tabcolsep{2pt}
\caption{Summary of the differences between our method and existing methods.}
\label{tab:intro_diff}
\begin{tabular}{@{}cccc@{}}
\toprule
Method           & \begin{tabular}[c]{@{}c@{}}Training-phase\\ Alteration-free\end{tabular} & \begin{tabular}[c]{@{}c@{}}Generation-phase\\ Alteration-free\end{tabular} & Model-agnostic \\ \midrule
Image Watermark~\cite{swanson1996transparent,luo2009reversible,pereira2000robust,tancik2020stegastamp}  & \textcolor{DarkGreen}{\ding{52}}                                               & \textcolor{red}{\ding{56}}                                                                           & \textcolor{DarkGreen}{\ding{52}}               \\
Classifier-based~\cite{yu2019attributing,ding2021does,yu2021artificial,yu2022responsible} &  \textcolor{red}{\ding{56}}                                                                        & \textcolor{DarkGreen}{\ding{52}}                                                                           & \textcolor{red}{\ding{56}}               \\
Ours             & \textcolor{DarkGreen}{\ding{52}}                                                                         & \textcolor{DarkGreen}{\ding{52}}                                                                           & \textcolor{DarkGreen}{\ding{52}}               \\ \bottomrule
\end{tabular}
\vspace{-0.3cm}
\end{wraptable}

In this paper, we propose a method for origin attribution that is based on the input reverse-engineering task on generative models (i.e., inverting the input of a particular generative model for a specific image). 
The intuition behind our method is that the reverse-engineering task is easier for belonging images than non-belonging images.
Therefore, we design our method based on the differences in the reconstruction loss for the reverse-engineering between generated images of a given model and other images.
The origin attribution method we propose starts by using the model to generate a set of images and calculate the reconstruction loss on these generated images.
To eliminate the influence of the inherent complexities of the examined images, we also calibrate the reconstruction loss by considering the hardness of the reverse-engineering on any other model that has strong generation ability, but
different architectures and training data.
Afterwards, the algorithm computes the calibrated reconstruction loss for the examined image and uses statistical hypothesis testing to determine if the reconstruction loss falls within the distribution of reconstruction losses observed in the generated images. This allows us to distinguish images generated by the given model from other images in a model-agnostic and alteration-free manner.
Based on our design, we implemented a prototype called \sys (\textbf{R}everse-engineering-based \textbf{O}rigi\textbf{N} \textbf{A}ttributio\textbf{N}) in PyTorch and evaluated it on three different types of generative models (i.e., unconditional model, class-to-image model, and text-to-image model) including various GANs~\cite{radford2015unsupervised,karras2020training,li2019controllable,zhang2018stackgan++,sauer2022stylegan}, VAEs~\cite{kingma2013auto}, and diffusion models such as latest Consistency Model~\cite{song2023consistency} and Stable Diffusion~\cite{rombach2022high}. Results demonstrate our method is effective for the ``alteration-free and model-agnostic origin attribution" task. On average, \sys achieves
95.70\% of true positive rate with a
false positive rate around 5.00\%.

Our contributions are summered as follows:
\ding{172} We introduce a new task called ``alteration-free and model-agnostic origin attribution", which entails determining whether a specific image is generated by a particular model without requiring any additional operations during the training and generation phases. \ding{173} To accomplish this task, we analyze the differences in the reconstruction loss for reverse-engineering between the generated images of a given model and other images. Based on our analysis, we design a novel method that involves conducting input reverse-engineering and checking whether the reconstruction loss of the examined sample falls within the distribution of reconstruction losses observed in the generated images.
\ding{174} We evaluate our method on eight different image generation models. The results show that our method effectively distinguishes images generated by the given model from other images in a model-agnostic and alteration-free manner.

\section{Related Work}\label{sec:related}

\noindent
\textbf{Detection of AI-Generated
Contents.} 
Detecting AI-generated content has become
extremely important with the growing concerns
about the misuse of AIGC
technology~\cite{chen2023pathway}. The detection of
AI-generated content is a binary classification problem
that involves distinguishing generated samples from real
ones. In the CV field, existing research has found that
visually imperceptible but machine-distinguishable patterns
in generated images, such as noise
patterns~\cite{zhao2021multi,marra2019gans}, frequency
signals~\cite{frank2020leveraging,durall2019unmasking,durall2020watch,jeong2022frepgan}
and texture representation~\cite{liu2020global} can be used
as the clues of AI-generated images. Researchers also
proposed methods to detect sentences generated by generative
NLP models such as ChatGPT~\cite{he2023mgtbench,chatgpt_classifier,jawahar2020automatic,galle2021unsupervised}.
Although these methods achieve promising performance for
distinguishing AI-generated content and real content, they
cannot infer if a specific content is generated by a given
generative model, which is a novel but more
challenging task and is the main focus of this paper.

\noindent
\textbf{Tracking Origin of Generated Images.}
There are several ways to track the source of the generated images.
Image watermarking that pastes ownership
information in carrier images~\cite{swanson1996transparent,luo2009reversible,pereira2000robust,tancik2020stegastamp} can be adapted to discern whether a specific sample is from a specific source.
Watermarks can take the form of specific signals within the images, such as frequency domain signals~\cite{pereira2000robust} or display-camera transformations~\cite{fang2018screen}.
However, it requires an additional modification to the generation results in a post-hoc manner, and it does not
necessarily reflect whether the image was generated by a particular model when the judges use it as the evidence (different models can use the same watermark). Another way is to inject fingerprints into the model during training and train a supervised classifier on it to discern whether an image is from a fingerprinted GAN model~\cite{yu2019attributing,ding2021does,yu2021artificial,yu2022responsible}. 
For example, Yu et al.~\cite{yu2022responsible} modify the architecture of the convolutional filter to embed the fingerprint, and they train the generator alongside a fingerprinting classifier capable of identifying the fingerprint and its corresponding source GAN models.
It requires a modified model architecture, altered training process, and an additional procedure to train a source classifier.

\section{Problem Formulation}\label{sec:problem}

We focus on serving as an inspector to infer \emph{if a specific sample is generated by a particular model in an alteration-free and model-agnostic manner}. To the best of our knowledge, this paper is the first work focusing on this problem.
To facilitate our discussion, we first define the belonging
and non-belonging of the generative models.

\begin{definition}[Belonging of Generative Models]\label{def:membership}
    Given a generative model \(\mathcal{M}: \mathcal{I} \mapsto \mathcal{X}_{\mathcal{M}}\) where \(\mathcal{I}\) is the input space and \(\mathcal{X}_{\mathcal{M}}\) is the output space. A sample \(\bm x\) is a \textbf{belonging} of model \(\mathcal{M}\) if and only if \(\bm x \in \mathcal{X}_{\mathcal{M}}\).
    We call a sample \(\bm x\) is a \textbf{non-belonging} if \(\bm x \notin \mathcal{X}_{\mathcal{M}}\).
\end{definition}

\noindent
\textbf{Inspector's Goal.}
Given a sample \(\bm x\) and a generative model \(\mathcal{M}: \mathcal{I} \mapsto \mathcal{X}_{\mathcal{M}}\) where \(\mathcal{I}\) is the input space and \(\mathcal{X}_{\mathcal{M}}\) is the output space of \(\mathcal{M}\),
the inspector's goal is to infer if a given image \(\bm x\) is a belonging of \(\mathcal{M}\).
Formally, the goal can be written as constructing an inference algorithm \(\mathcal B: (\mathcal{M},\bm x)
    \mapsto \{0,1\}\) that receives a sample \(\bm x\) and a model \(\mathcal{M}\) as the input, and returns the inference result (i.e., 0 denotes belongings, and 1 denotes non-belongings). The algorithm \(\mathcal B\) can distinguish not only the belongings of the given model \(\mathcal{M}\) and that of the other models (e.g., trained on different training data, or have different model architectures), but also the belongings of \(\mathcal{M}\) and the natural samples that are not generated by AI.
The inspector also aims to achieve the following goals:

\emph{Alteration-free:} The algorithm \(\mathcal B\) does not require any extra operations/modifications in the training phase and image generation phase.

\emph{Model-agnostic:} The algorithm \(\mathcal B\) can be applied to different types of image generative models with different architectures.

\noindent
\textbf{Inspector's Capability.}
The inspector has white-box access to the provided model \(\mathcal{M}\), thus the inspector can get the intermediate output and calculate the gradient of the models.
In addition, the inspector cannot control the development and training process of the provided models.

\noindent
\textbf{Real-world Application.} 
The inspection algorithm can be widely used in various scenarios where it is necessary to verify the authenticity of generated images. We provide three examples as follows:

\emph{Copyright protection of image generation models:} %
A copyright is a kind of intellectual property (IP) that provides its owner the exclusive right to perform a creative work~\cite{spence2007intellectual}.
In this scenario, a party suspects that a specific image may have been generated by their generative model without authorization, such as if a malicious user has stolen the model and used it to generate images. The party can then request an inspector to use our proposed method to infer if the doubtful image was indeed generated by their particular model, and the resulting inference can be used as a reference in a lawsuit. It is important to note that %
this would be specifically useful for cases where the copyright of the generated images should belong to the model owner%
, rather than the user who provides the prompts.
The main goal is to safeguard the copyright of the trained model and the generated images.

\emph{Tracing the source of maliciously generated images:} 
Assume a user creates malicious images containing inappropriate or biased content and distributes them on the internet. The cyber police can utilize our proposed method to infer if the image was generated by a model belonging to a specific user. The resulting inference can be used as a reference for criminal evidence in lawsuits.

\emph{Detecting AI-powered plagiarism:}
Assume a user creates AI-generated images and falsely claims them as their own artwork in an artwork competition or exam. The inspector can utilize our proposed method to detect if the submitted images were created by open-sourced image generation models. This can help ensure fairness and protect other creators' rights in competitions and exams.

\section{Method}\label{sec:method}

Our method is built on the input reverse-engineering for image generation models.
In this section, we start by formulating the input reverse-engineering task, followed by an analysis of the disparities in reconstruction loss between images generated by a particular model and those from other sources. We then proceed to present a detailed algorithm for our method.

\subsection{Reverse-engineering}

We view the reverse-engineering as an optimization problem. 
Formally, it can be defined as follows:

\begin{definition}[Input Reverse-engineering]\label{def:reconstruction}
    Given a generative model \(\mathcal{M}: \mathcal{I}
    \mapsto \mathcal{X}_{\mathcal{M}}\), and an image \(\bm
    x\), an input reverse-engineering task is optimizing the input \(\bm i\) to make the corresponding output image from the model \(\mathcal{M}(\bm i)\) as close as possible to the given image \(\bm x\).
\end{definition}

The input reverse-engineering task is performed by a \textbf{reverse-engineering algorithm} 
\(\mathcal{A}: (\mathcal{M},\bm x)
    \mapsto R\), which can be written as:

\vspace{-0.4cm}
\begin{equation}\label{eq:re}
    \bm i^{\star} = \argmin\limits_{\bm i}
    \mathcal{L}\left(\mathcal{M}(\bm i), \bm x\right),\quad \mathcal{A}(\mathcal{M}, \bm x) = \mathcal{L}\left(\mathcal{M}(\bm i^{\star}), \bm
x\right)
\end{equation}
where \(\mathcal{L}\) is a metric to measure the
distance between different images, and \(\bm i^{\star}\) is the reverse-engineered input.
The given value for the reverse-engineering algorithm \(\mathcal{A}\) is a specific image \(\bm{x}\) and a particular model \(\mathcal{M}\). The returned
value of the algorithm \(\mathcal{A}\) is the distance between the given image and its reverse-engineered version, i.e., \(\mathcal{L}\left(\mathcal{M}(\bm i^{\star}), \bm
x\right)\), which is called as the \textbf{reconstruction loss}. We use the reconstruction loss to measure the hardness of the input reverse-engineering task.

Based on the definitions and formulations, we have the following theorem:

\begin{theorem}\label{th:theo1}
    Given a generative model \(\mathcal{M}: \mathcal{I}
    \mapsto \mathcal{X}_{\mathcal{M}}\), and a
    reverse-engineering algorithm \(\mathcal{A}\), if the model
    is deterministic (i.e., it produces the same
    output given the same input) and the reverse-engineering
    algorithm is perfect (i.e., it can find the global minimum of the
    reconstruction loss for the reverse-engineering), then for any \(\bm x \in
    \mathcal{X}_{\mathcal{M}}\) (belonging) and \(\bm
    x^{\prime} \notin \mathcal{X}_{\mathcal{M}}\)
    (non-belonging) we have \(\mathcal{A}(\mathcal{M}, \bm x^{\prime}) >
    \mathcal{A}(\mathcal{M}, \bm x)\).
\end{theorem}

The proof for \autoref{th:theo1} can be found in
the Appendix. The theorem demonstrates that the reconstruction loss of the images generated by a specific model will be lower than that of images that are not generated by the model. The theorem also establishes that the distribution of reconstruction loss values for belonging and non-belonging images is perfectly separable. Thus, we can use a
threshold value to separate the belonging images and non-belonging images. 
In the real world, many image generation models incorporate random noises into their image generation procedures to enhance the variety of images they produce. However, these models can also be deemed deterministic since we can regard all the random noises utilized in the generation procedure as parts of the inputs.
On the other hand, in reality, the reverse-engineering algorithm may get stuck at a local minimum, and it is hard to guarantee the achievement of the global minimum. This is where 
the formula \(\mathbb{P}(\mathcal{A}(\mathcal{M}, \bm x^{\prime}) >\mathcal{A}(\mathcal{M}, \bm x))\geq \lambda\)
becomes relevant as it serves as a relaxation for \autoref{th:theo1}, explaining the practical scenario. In this formula, 
\(\lambda\) (e.g., \(90\%\)) acts as a separability level for distinguishing between the two distributions: belonging images and non-belonging images.

\begin{figure}[]
    \begin{subfigure}[t]{0.33\columnwidth}
        \centering
        \includegraphics[width=\columnwidth]{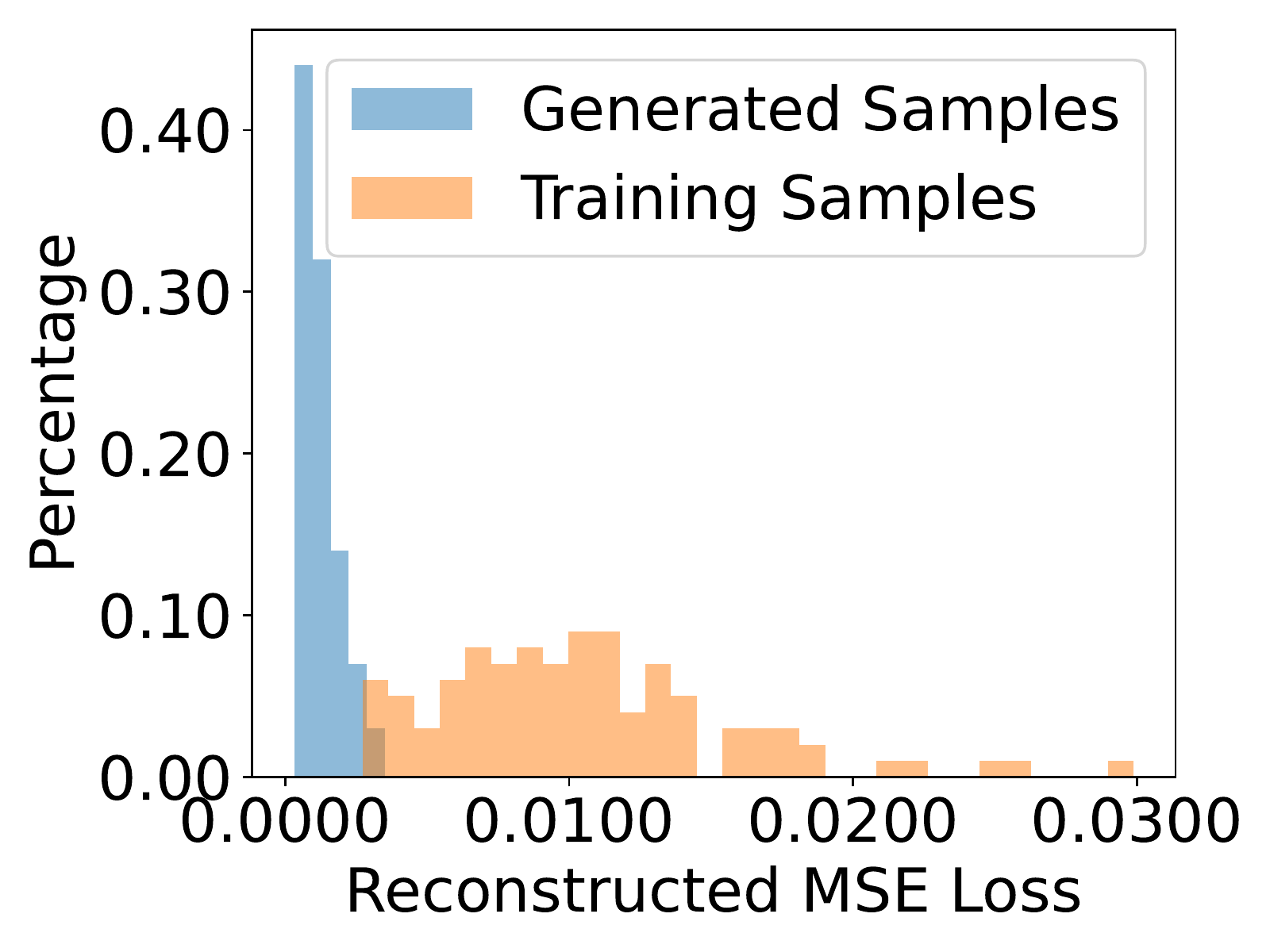}
        \caption{DCGAN~\cite{radford2015unsupervised}}
    \end{subfigure}
    \begin{subfigure}[t]{0.33\columnwidth}
        \centering
        \includegraphics[width=\columnwidth]{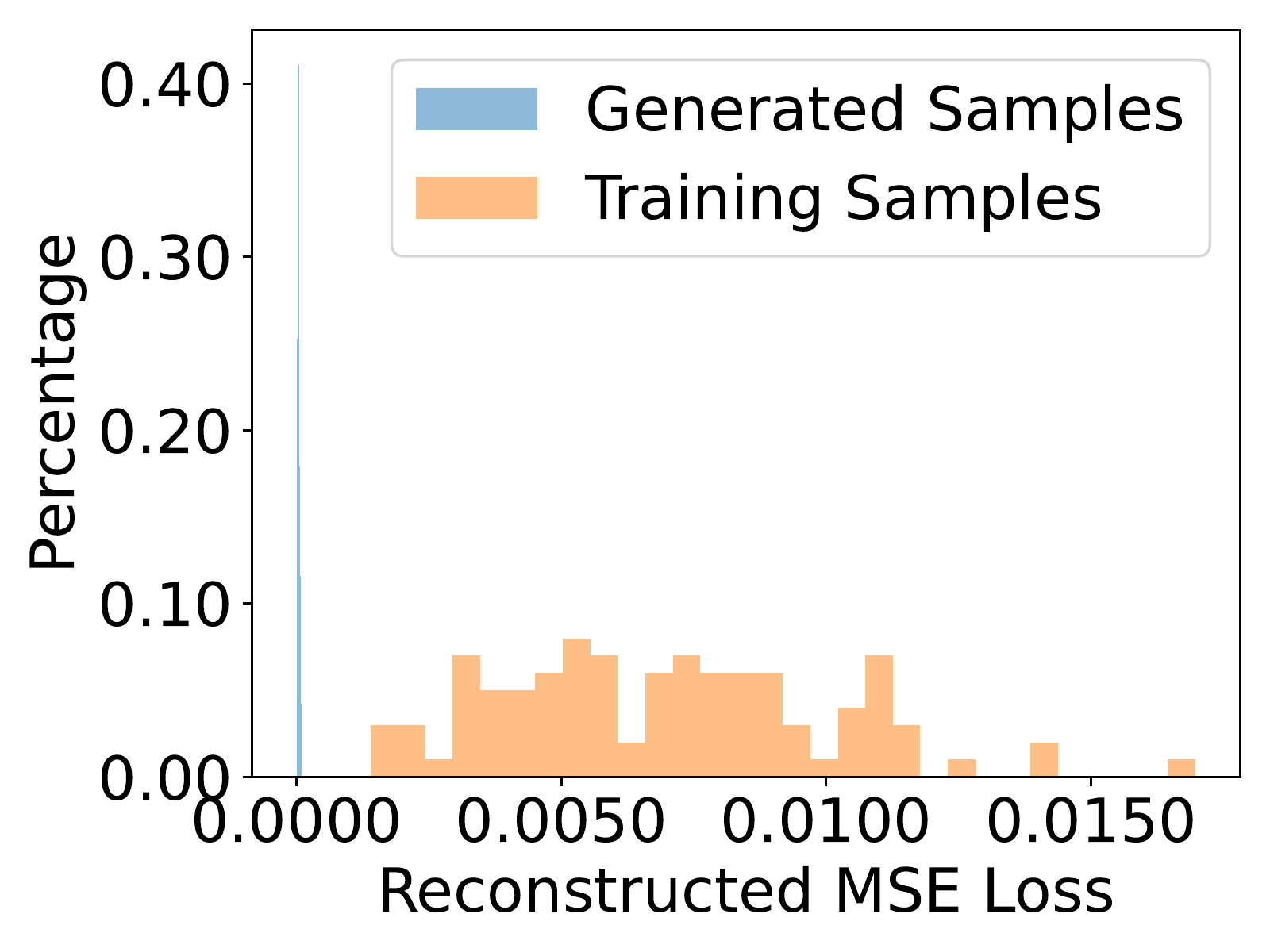}
        \caption{VAE~\cite{kingma2013auto}}
    \end{subfigure}
    \begin{subfigure}[t]{0.33\columnwidth}
        \centering
        \includegraphics[width=\columnwidth]{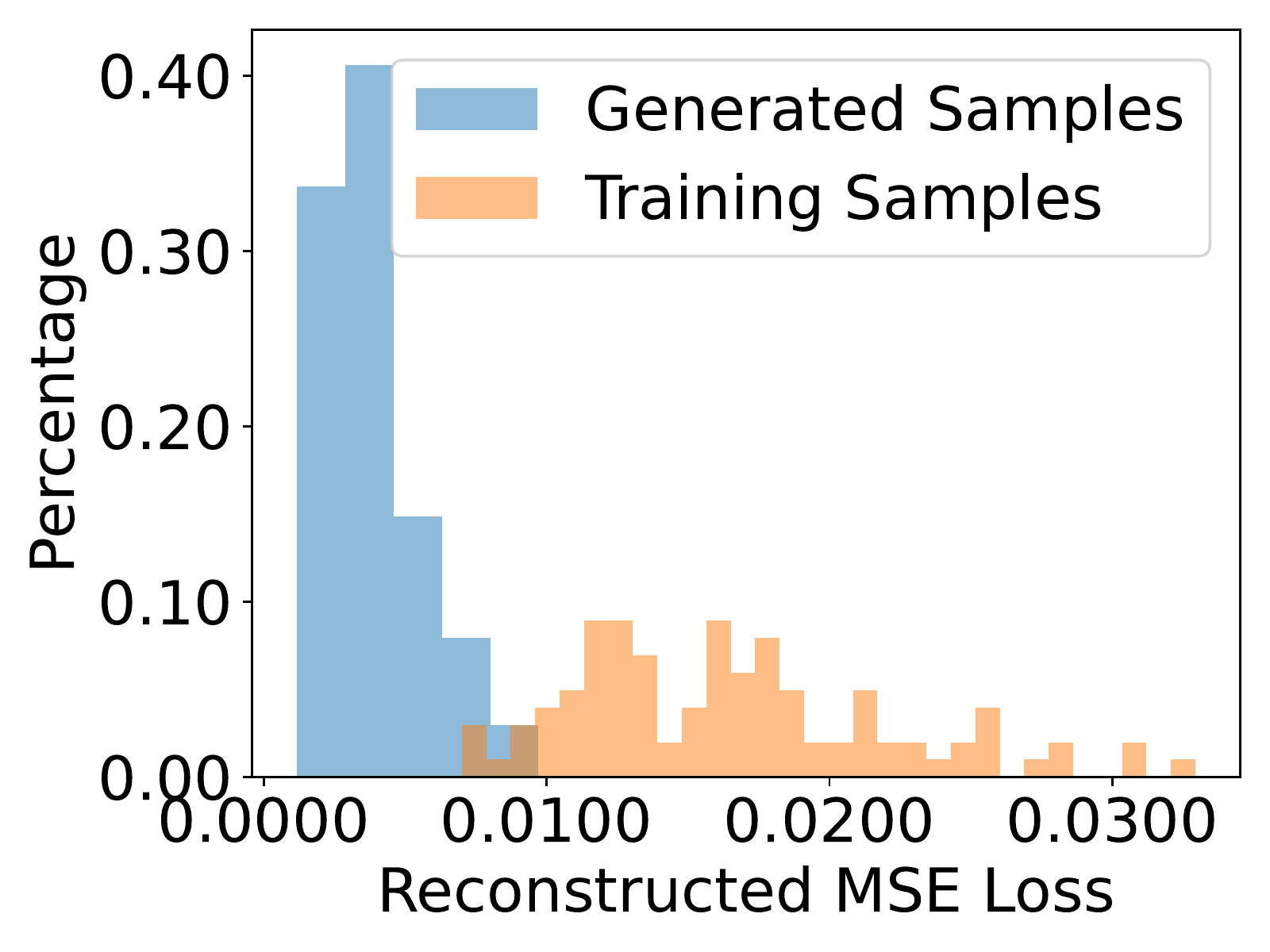}
        \caption{StyleGAN2-ADA~\cite{karras2020training}}
    \end{subfigure}
    \caption{Reconstruction loss distributions for belonging images and real images.}
    \label{fig:observation}
    \vspace{-0.3cm}
\end{figure}

To investigate the practical scenario, we conduct experiments on the CIFAR-10~\cite{krizhevsky2009learning} dataset using DCGAN~\cite{radford2015unsupervised}, VAE~\cite{kingma2013auto}, and StyleGAN2-ADA~\cite{karras2020training} models. The results are depicted in \autoref{fig:observation}, where the x-axis represents the reconstruction loss measured by the MSE (Mean Squared Error)~\cite{wang2009mean} metric, and the y-axis indicates the percentage of images whose reconstruction loss value corresponds to the corresponding value on the x-axis. We use blue color to denote 100 generated images of the given model and orange to represent 100 real images randomly sampled from the training data of the model. The results indicate that the reconstruction losses of the generated images (belongings) and those not generated by this model (non-belongings) can be distinguished.

\subsection{Calibration}
\label{sec:cali}
Different images have different inherent complexities~\cite{redies2012phog,yu2013image,mario2005image}.
Some images may be harder to reverse-engineer due to their higher complexity (e.g., containing more objects, colors, and details). In that case, the reconstruction loss will also be influenced by the inherent complexities of the examined images. To increase the separability level of belonging images and others, we disentangle the influence of the inherent complexities and the belonging status by using a reference image generation model \(\mathcal{M}_r\) that is trained on a different dataset (we use the Consistency model~\cite{song2023consistency} pre-trained on ImageNet dataset~\cite{russakovsky2015imagenet} as the reference model by default). The calibrated reconstruction loss \(\mathcal{A}^{\prime}(\mathcal{M}, \bm x)\) is defined as follows:

\vspace{-0.3cm}
\begin{equation}\label{eq:cali}
    \mathcal{A}^{\prime}(\mathcal{M}, \bm x) = \frac{\mathcal{A}(\mathcal{M}, \bm x)}{\mathcal{A}(\mathcal{M}_r, \bm x)}
\end{equation}
\vspace{-0.4cm}

\subsection{Belonging Inference via Hypothesis Testing}
\label{sec:hypo}
We use Grubbs' Hypothesis Testing~\cite{grubbs1950sample} to infer if a specific sample \(\bm x\) is a belonging of the particular given model \(\mathcal M\).
We have a null hypothesis \(H_0:\) \emph{\(\bm x\) is a non-belonging of \(\mathcal M\)}, and the alternative hypothesis \(H_1:\) \emph{\(\bm x\) is a belonging of \(\mathcal M\)}. The null hypothesis \(H_0\) is rejected (i.e., the alternative hypothesis \(H_1\) is accepted) if the following inequality (\autoref{eq:hypo}) holds:

\begin{equation}
\label{eq:hypo}
\frac{\mathcal{A}^{\prime}(\mathcal{M}, \bm x) - \mu}{\sigma} <\frac{(N-1)}{\sqrt{N}} \sqrt{\frac{\left(t_{\alpha /N, N-2}\right)^2}{N-2+\left(t_{\alpha /N, N-2}\right)^2}}
\end{equation}

Here, \(\mu\) and \(\sigma\) are the mean value and standard deviation for the calibrated reconstruction loss on belonging samples of model \(\mathcal{M}\). Since model \(\mathcal{M}\) is given to the inspector, the inspector can calculate \(\mu\) and \(\sigma\) by using \(\mathcal{M}\) to generate multiple images with randomly sampled inputs. \(N\) is the number of generated belonging images.
\(\mathcal{A}^{\prime}(\mathcal{M}, \bm x)\) is the calibrated reconstruction loss of the examined image \(\bm x\). \(t_{\alpha /N, N-2}\) is the critical value of the \(t\) distribution with  \(N-2\) degrees of freedom and a significance level of \(\alpha /N\), where \(\alpha\) is the significance level of the hypothesis testing (i.e., 0.05 by default in this
paper). The critical value of the \(t\) distribution (i.e., \(t_{\alpha /N, N-2}\)) can be computed using the cumulative distribution function (See Appendix for more details).%

\begin{algorithm}[tb]
 	\caption{Origin Attribution}\label{alg:detection}
    {\bf Input:} %
    \hspace*{0.05in} Model: \(\mathcal{M}\), Examined Data: \(\bm x\)\\
    {\bf Output:} %
    \hspace*{0.05in} Inference Results: Belonging or Non-belonging
	\begin{algorithmic}[1]
	     \Function {Inference}{$\mathcal{M}, \bm x$}
      \LineComment{Obtaining Belonging Distribution (Offline)}
      \State \(\mu, \sigma, N = {\rm BelongingDistribution}(\mathcal{M})\)%
      \LineComment{Reverse-engineering}
      \State \(\mathcal{A}^{\prime}(\mathcal{M}, \bm x) \leftarrow\) Calibrated Reconstruction Loss [\autoref{eq:cali}]

      \LineComment{Determining Belonging}
      \State \({\rm InferenceResults = \rm HypothesisTesting}(\mathcal{A}^{\prime}(\mathcal{M}, \bm x),\mu, \sigma, N)\)[\autoref{eq:hypo}]
      \State \Return{$\rm InferenceResults$}
    \EndFunction
    \end{algorithmic}
\end{algorithm}

\subsection{Algorithm}

We propose \autoref{alg:detection} to determine if a specific sample belongs to a given model. The input of \autoref{alg:detection} is the examined data 
\(\bm x\) and the given model 
\(\mathcal{M}\). The output of this algorithm is the inference results, i.e., belonging or non-belonging. In line 3, we use the given model to generate \(N\) (i.e., 100 by default in this paper) images with randomly sampled inputs and calculate the mean value (\(\mu\)) and standard deviation (\(\sigma\)) for the calibrated reconstruction loss on the generated belonging samples.
This step can be done offline, i.e., it only needs to be performed once for each model.
In line 5, we calculate the calibrated reconstruction loss of the examined image (\autoref{eq:cali}), the reconstruction loss is computed via gradient descent optimizer (Adam~\cite{kingma2014adam} by default in this paper).
In line 7, we determine if the examined image 
\(\bm x\) belongs to the model 
\(\mathcal{M}\) or not by conducting the Grubbs' Hypothesis Testing~\cite{grubbs1950sample} (\autoref{sec:hypo}). The given image is flagged as a belonging of the given model if the corresponding hypothesis is accepted.

\section{Experiments and Results}
\label{sec:eval}

In this section, we first introduce the setup of the experiments (\autoref{sec:setup}). We
evaluate the effectiveness of \sys (\autoref{sec:effe}) and provide a case study on Stable Diffusion 2.0 model~\cite{rombach2022high} (\autoref{sec:sd}). We then conduct ablation studies in \autoref{sec:ablation}. The discussion about the efficiency and robustness against image editing can be found in the Appendix.

\subsection{Setup}
\label{sec:setup}
Our method is implemented with Python 3.8 and PyTorch 1.11.
We conducted all experiments on a Ubuntu 20.04 server equipped with 64 CPUs and six Quadro RTX 6000 GPUs.

\noindent
\textbf{Models.}
Eight different models are included in the experiments:
DCGAN~\citep{radford2015unsupervised}, VAE~\citep{kingma2013auto}, StyleGAN2-ADA~\citep{karras2020training}, StyleGAN XL~\cite{sauer2022stylegan}, Consistency Diffusion Model~\citep{song2023consistency}, Control-GAN~\citep{li2019controllable},
StackGAN-v2~\cite{zhang2018stackgan++},
and Stable Diffusion 2.0~\citep{rombach2022high}. These models are representative image generation models. 

\noindent
\textbf{Performance Metrics.}
The effectiveness of the method is measured by collecting the detection accuracy (Acc).
For a particular model, given a set of belonging images and non-belonging images, the Acc is the ratio between the correctly classified images and all images.
We also show a detailed number of True Positives (TP, i.e., correctly detected belongings), False Positives (FP, i.e., non-belongings classified as belongings), False Negatives (FN, i.e., belongings classified as non-belongings) and True Negatives (TN, i.e., correctly classified non-belongings).

\subsection{Effectiveness}
\label{sec:effe}
In this section, we evaluate the effectiveness of \sys%
from two perspectives: %
(1) its effectiveness in distinguishing between belongings of a particular model and real images; %
(2) its effectiveness in differentiating between belongings of a particular model and those generated by other models.

\noindent
\textbf{Distinguishing Belonging Images and Real Images.}
To investigate \sys's effectiveness in distinguishing between belonging images of a particular model and real images, given an image generation model, we start by differentiating between the generated images of the given model and the training data of the model. The investigated models %
include DCGAN~\cite{radford2015unsupervised}, VAE~\cite{kingma2013auto}, StyleGAN2-ADA~\cite{karras2020training} trained on the CIFAR-10~\cite{krizhevsky2009learning} dataset, Consistency Model~\cite{song2023consistency} trained on the ImageNet~\cite{russakovsky2015imagenet} dataset, and ControlGAN~\cite{li2019controllable} trained on the CUB-200-2011~\cite{WahCUB_200_2011} dataset.
Among them, DCGAN and VAE are unconditional image generation models. StyleGAN2-ADA and the latest diffusion model Consistency Model, are class-conditional models.
In addition to distinguishing belongings and the training data, we also conduct experiments to distinguish belongings from unseen data that has a similar distribution to the training data (i.e., the test data of the dataset). 
For each case, we evaluate the results on 100 randomly sampled belonging images and 100 randomly sampled non-belonging images.
The results are demonstrated in \autoref{tab:unconditional}. As can be observed, the detection accuracies (Acc) are above 85\% in all cases. On average, the Acc is 94.2\% for distinguishing belongings and training data, and is 95.9\% for distinguishing belongings and unseen data. The results demonstrate that \sys achieves good performance in distinguishing between belonging images of a particular model and real images.

\begin{table}[]
\centering
\scriptsize
\setlength\tabcolsep{3pt}
\caption{Detailed results on distinguishing belonging images and real images.}\label{tab:unconditional}
\begin{tabular}{@{}cccccccccccccc@{}}
\toprule
\multirow{2}{*}{Model Type}        & \multirow{2}{*}{Model} & \multirow{2}{*}{Training Dataset} & \multicolumn{5}{c}{Belongings vs Training Data} &  & \multicolumn{5}{c}{Belongings vs Unseen Data} \\ \cmidrule(lr){4-8} \cmidrule(l){10-14} 
                                   &                        &                                   & TP     & FP     & FN     & TN      & Acc       &  & TP     & FP     & FN    & TN     & Acc       \\ \midrule
\multirow{2}{*}{Unconditional}     & DCGAN                  & CIFAR-10                          & 96     & 0      & 4      & 100     & 98.0\%    &  & 95     & 0      & 5     & 100    & 97.5\%    \\
                                   & VAE                    & CIFAR-10                          & 95     & 0      & 5      & 100     & 97.5\%    &  & 96     & 0      & 4     & 100    & 98.0\%    \\ \midrule
\multirow{2}{*}{Class-conditional} & StyleGAN2-ADA          & CIFAR-10                          & 96     & 2      & 4      & 98      & 97.0\%    &  & 95     & 0      & 5     & 100    & 97.5\%    \\
                                   & Consistency Model      & ImageNet                          & 96     & 24     & 4      & 76      & 86.0\%    &  & 96     & 11     & 4     & 89     & 92.5\%    \\ \midrule
Text-conditional                   & ControlGAN             & CUB-200-2011                      & 95     & 10     & 5      & 90      & 92.5\%    &  & 96     & 8      & 4     & 92     & 94.0\%    \\ \bottomrule
\end{tabular}
\end{table}
\begin{table}[]
\centering
\scriptsize
\caption{Results for distinguishing belonging images and images generated by other models with different architectures. Here, Model \(\mathcal{M}_1\) is the examined model, Model \(\mathcal{M}_2\) is the other model that has same training data but different architectures.}\label{tab:different_arch}
\begin{tabular}{@{}cccccccc@{}}
\toprule
Training Dataset              & Model \(\mathcal{M}_1\)           & Model \(\mathcal{M}_2\)           & TP & FP & FN & TN  & Acc    \\ \midrule
\multirow{6}{*}{CIFAR-10}     & DCGAN             & VAE               & 96 & 0  & 4  & 100 & 98.0\% \\
                              & VAE               & DCGAN             & 97 & 0  & 3  & 100 & 98.5\% \\
                              & DCGAN             & StyleGAN2ADA      & 96 & 0  & 4  & 100 & 98.0\% \\
                              & StyleGAN2ADA      & DCGAN             & 96 & 2  & 4  & 98  & 97.0\% \\
                              & VAE               & StyleGAN2ADA      & 95 & 0  & 5  & 100 & 97.5\% \\
                              & StyleGAN2ADA      & VAE               & 95 & 1  & 5  & 99  & 97.0\% \\ \midrule
\multirow{2}{*}{ImageNet}     & Consistency Model & StyleGAN XL       & 95 & 8  & 5  & 92  & 93.5\% \\
                              & StyleGAN XL       & Consistency Model & 96 & 10 & 4  & 90  & 93.0\% \\ \midrule
\multirow{2}{*}{CUB-200-2011} & ControlGAN        & StackGAN-v2       & 96 & 17 & 4  & 83  & 89.5\% \\
                              & StackGAN-v2       & ControlGAN        & 96 & 14 & 4  & 86  & 91.0\% \\ \bottomrule
\end{tabular}
\end{table}

\noindent
\textbf{Distinguishing Belonging Images and Images Generated by Other Models.}
In this section, we study \sys's effectiveness to distinguish between belonging images of a particular model and the images generated by other models. For a given model \(\mathcal{M}_1\), we consider two different types of other models \(\mathcal{M}_2\) 
, i.e., the model trained on the same dataset but with different architectures and the model that has the same architecture but is trained on a different dataset.

\emph{Models with Different Architectures.}
We first evaluate \sys's effectiveness in distinguishing between belonging images of a particular model and the images generated by other models with the same training data but different architectures. For the model trained on the CIFAR-10~\cite{krizhevsky2009learning} dataset, the used model architectures are 
DCGAN~\cite{radford2015unsupervised}, VAE~\cite{kingma2013auto}, StyleGAN2-ADA~\cite{karras2020training}. For the Imagenet~\cite{russakovsky2015imagenet} dataset, the involved models are the latest diffusion model Consistency Model~\cite{song2023consistency} and StyleGAN XL~\cite{sauer2022stylegan}.
For the CUB-200-2011~\cite{WahCUB_200_2011} dataset, we use text-to-image models 
ControlGAN~\cite{li2019controllable}
and StackGAN-v2~\cite{zhang2018stackgan++}.
To measure the effectiveness of \sys,
we collect its results on 100 randomly sampled belonging images and 100 randomly sampled non-belonging images. The results are shown in \autoref{tab:different_arch}, where Model \(\mathcal{M}_1\) denotes the examined model, and Model \(\mathcal{M}_2\) represents the other model that has the same training data but a different architecture.
The results show that the average detection accuracy (Acc) of \sys is 95.3\%, confirming its good performance for distinguishing between belongings of a given model and the images generated by other models with different architectures.

\emph{Models Trained on Different %
Datasets.}
We also evaluate \sys's effectiveness in distinguishing between belonging images of a 
particular model and the images generated by other models with the same model architecture but trained on different %
datasets.
The model used here is the diffusion model Consistency Model~\cite{song2023consistency}. We use a model trained on 
the ImageNet~\cite{russakovsky2015imagenet} dataset and a model trained on the LSUN~\cite{yu2015lsun} dataset. The results are demonstrated in \autoref{tab:different_trainingdata},
\begin{wraptable}{r}{0.5\linewidth}
\centering
\scriptsize
\setlength\tabcolsep{3pt}
\caption{Results for distinguishing belonging images and images generated by other models with different training data. Here, Model \(\mathcal{M}_1\) is the examined model, Model \(\mathcal{M}_2\) is the other model that has same architecture but different training data.}\label{tab:different_trainingdata}
\begin{tabular}{@{}ccccccc@{}}
\toprule
\begin{tabular}[c]{@{}c@{}}Trainging dataset \\ of Model \(\mathcal{M}_1\)\end{tabular} & \begin{tabular}[c]{@{}c@{}}Trainging dataset \\ of Model \(\mathcal{M}_2\)\end{tabular} & TP & FP & FN & TN & Acc    \\ \midrule
ImageNet                                                                & LSUN                                                                    & 95 & 13 & 5  & 87 & 91.0\% \\
LSUN                                                                    & ImageNet                                                                & 96 & 6  & 4  & 94 & 95.0\% \\ \bottomrule
\end{tabular}
\end{wraptable}
where Model \(\mathcal{M}_1\) denotes the examined model, and Model \(\mathcal{M}_2\) means the other model that has the same model architecture but is trained on a different dataset. The results show that \sys can effectively distinguish between belonging images of a particular model and the images generated by other models with the same model architecture but different training data. On average, the detection accuracy of our method is 93.0\%. In the Appendix, we also discuss the results when the training data of the model \(\mathcal{M}_2\) and that of the model \(\mathcal{M}_1\) have overlaps. Empirical results demonstrate that \sys is still effective even the training data of the examined model and that of the other model are similar (i.e., they have large overlaps).

\begin{figure*}[b]
    \captionsetup[subfigure]{labelformat=empty}
    \centering
    \subfloat[(a) A non-belonging image and its reverse-engineered image for Stable Diffusion 2.0. The reconstructed loss is 0.0078 with MSE metric.]
    {
    \subfloat[Non-belonging]{\includegraphics[width=0.22\columnwidth]{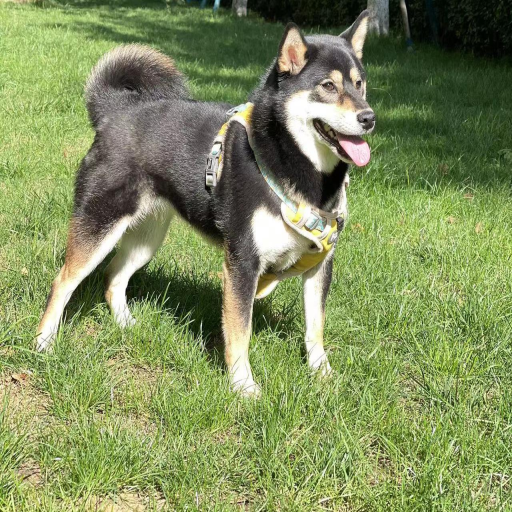}}
    \hspace{0.1cm}
    \subfloat[Reverse-engineered]{\includegraphics[width=0.22\columnwidth]{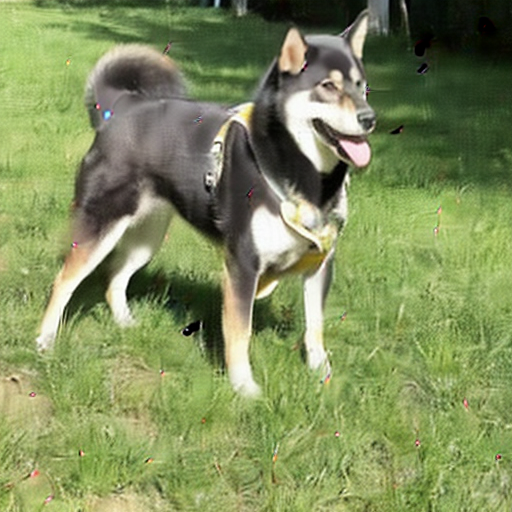}}
    }
    \hfill
    \subfloat[(b) A belonging image and its reverse-engineered image for Stable Diffusion 2.0. The reconstructed loss is 0.0005 with MSE metric.]
    {\setcounter{subfigure}{0}
    \subfloat[Belonging]{\includegraphics[width=0.22\columnwidth]{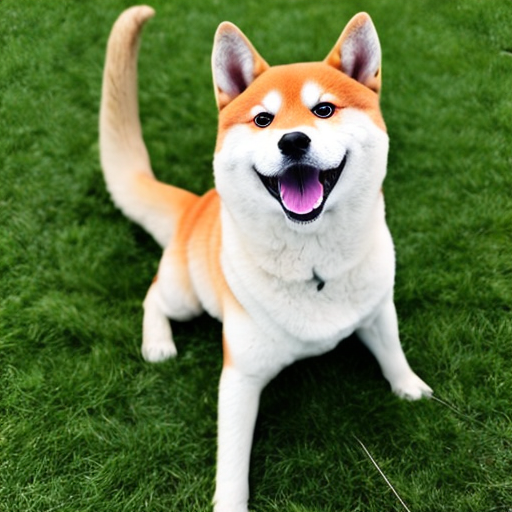}}
    \hspace{0.1cm}
    \subfloat[Reverse-engineered]{\includegraphics[width=0.22\columnwidth]{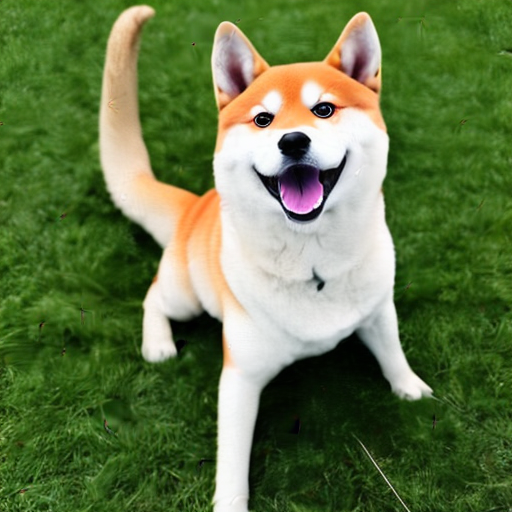}}
    }
    \hfill
    \caption{Visualization of the belonging image and non-belonging image for Stable Diffusion 2.0~\cite{rombach2022high}, and their corresponding reverse-engineered images. The non-belonging image and the belonging image have similar inherent complexities.
    }\label{fig:visualization_stable_diffusion}
\end{figure*}

\subsection{Case Study on Stable Diffusion 2}\label{sec:sd}
In this section, we conduct a case study on the recent Stable Diffusion 2.0~\cite{rombach2022high} model. We first randomly collect 20 images 
of Shiba dogs from the internet and use these images as the non-belonging images. We then use the prompt "A cute Shiba on the grass." and feed it into the Stable Diffusion 2.0 model to generate 20 belonging images. We apply \sys on the model, and evaluate its performance in distinguishing the belonging images and non-belonging images. 
The results show that the detection accuracy of \sys is 87.5\%, with 18 TP, 3 FP, 2 FN, 17 TN. In \autoref{fig:visualization_stable_diffusion}, we show the visualization of a belonging image and a non-belonging image, as well as their corresponding reverse-engineered images. Note that the non-belonging image and the belonging image have similar inherent complexities (i.e., their reconstructed loss with the MSE metric on the reference model are 0.029 and 0.022, respectively). 
For the non-belonging image, the reverse-engineered image is more noisy and blurred, while the reverse-engineered image of the belonging image seems nearly identical to the original image. 
These results show the potential to apply our method on state-of-the-art  models such as Stable Diffusion 2.0.
More visualizations and examples can be found in the Appendix.

\subsection{Ablation Study}
\label{sec:ablation}
In this section, we evaluate the impact of different metrics used in calculating reconstruction loss, and the impact of reconstruction loss calibration.

\noindent
\textbf{Different Metrics.}
In the reverse-engineering task (\autoref{eq:re}), we use a metric \(\mathcal{L}\) to measure the distance between different images. 
By default, we select MSE~\cite{wang2009mean} as the metric.
\begin{wraptable}{r}{0.45\linewidth}
\vspace{-0.3cm}
\centering
\scriptsize
\setlength\tabcolsep{3pt}
\caption{Results on different distance metrics.}
\label{tab:metrics}
\begin{tabular}{@{}cccccc@{}}
\toprule
Metric & TP & FP & FN & TN & Acc    \\ \midrule
MAE                 & 96 & 7  & 4  & 93 & 94.5\% \\
MSE                 & 96 & 2  & 4  & 98 & 97.0\% \\
SSIM                & 97 & 26 & 3  & 74 & 85.5\% \\
LPIPS               & 96 & 4  & 4  & 96 & 96.0\% \\ \bottomrule
\end{tabular}
\vspace{-0.3cm}
\end{wraptable}

In addition to MSE, we also evaluate the results on other image distance metrics, i.e., MAE (Mean Absolute Error)~\cite{chai2014root}, SSIM (Structural Similarity Index Measure)~\cite{wang2004image}, and LPIPS (Learned Perceptual Image Patch Similarity)~\cite{zhang2018unreasonable}. The task %
is to distinguish between belonging images and real images (i.e., training images of the model here). The model used in this section is the StyleGAN2-ADA~\cite{karras2020training} trained on CIFAR-10~\cite{krizhevsky2009learning} dataset. The results are shown in \autoref{tab:metrics}.
As we can observe, the detection accuracy (Acc) under MAE, MSE, SSIM, and LPIPS are 94.5\%, 97.0\%, 85.5\%, and 96.0\%, respectively. Overall, the MSE metric achieves the highest performance in distinguishing belonging images and real images. Thus, we select MSE as our default metric.

\noindent
\textbf{Impacts of Reconstruction Loss Calibration.} To eliminate the influence of images' inherent complexities, we calibrate the reconstruction loss by considering the hardness \begin{wraptable}{r}{0.49\linewidth}
\vspace{-0.3cm}
\centering
\scriptsize
\setlength\tabcolsep{3pt}
\caption{Effects of reconstruction loss calibration.}\label{tab:cali}
\begin{tabular}{@{}cccccc@{}}
\toprule
Method          & TP & FP & FN & TN & Acc    \\ \midrule
w/o Calibration & 17 & 7  & 3  & 13 & 75.0\% \\
w/ Calibration  & 18 & 3  & 2  & 17 & 87.5\% \\ \bottomrule
\end{tabular}
\vspace{-0.3cm}
\end{wraptable}
 of the reverse-engineering on a reference model (\autoref{sec:cali}). To measure the effects of the calibration step, we compare the detailed TP, FP, FN, TN, and Acc for the method with and without the calibration step. We use the Stable Diffusion 2.0~\cite{rombach2022high} model and follow the experiment settings described in \autoref{sec:sd}. The results 
in \autoref{tab:cali} demonstrate the detection accuracy for the method with and without the calibration step are 75.0\% and 87.5\%, respectively. These results show that the calibration step can effectively 
improve the performance of \sys.

\section{Discussion}\label{sec:discussion}

\textbf{Limitations.}
While our method can achieve origin attribution in a alteration-free and model-agnostic manner, the computation cost might be higher than that of watermark-based methods~\cite{swanson1996transparent,luo2009reversible,pereira2000robust,tancik2020stegastamp} and classifier-based methods~\cite{yu2019attributing,ding2021does,yu2021artificial,yu2022responsible}. More discussion about the efficiency of our method can be found in the Appendix. Speeding up the reverse-engineering will be our future work.
This paper focuses on the image generation models. Expanding our method for origin attribution on the model in other fields, (e.g., vedio generation models such as Imagen Video~\cite{ho2022imagen}, language generation models like ChatGPT~\cite{chatgpt}, and graph generation models~\cite{xu2022geodiff,vignac2022digress}) will also be our future work. 

\textbf{Ethics.}
Research on security and privacy of machine learning potentially has ethical concerns~\cite{tramer2016stealing,wang2023unicorn,liu2022complex,mei2023notable,shokri2017membership,wang2022training,cheng2018query}.
This paper proposes a method to safeguard the intellectual property of image generation models and monitor their potential misuse. We believe that our approach will enhance the security of image generation models and be beneficial to generative AI.

\section{Conclusion}
\label{sec:conclusion}

In this paper, we take the first effort to introduce the ``alteration-free and model-agnostic origin attribution'' task for AI-generated images.
Our contributions for accomplishing this task involves defining the reverse-engineering task for generative models and analyzing the disparities in reconstruction loss between generated samples of a given model and other images. Based on our analysis, we devise a novel method for this task by conducting input reverse-engineering and calculating the corresponding reconstruction loss. Experiments conducted on different generative models under various settings demonstrate that our method is effective
in tracing the origin of AI-generated images.

\bibliographystyle{unsrtnat}
\bibliography{reference}

\newpage

{\bf Roadmap:} In this appendix, %
we show the proof for Theorem 4.2 (\autoref{sec:proof}),
detailed process for calculating critical value of the \(t\) distribution used in \autoref{sec:hypo} (\autoref{sec:appendix_critical}).
more results for distinguishing belonging images and images generated by other models (\autoref{sec:appendix_trainingdata}), \sys's robustness to editing-based adaptive attack (\autoref{sec:appendix_editing}), the discussion for the efficiency of \sys (\autoref{sec:appendix_efficiency}). Finally, we demonstrate more visualizations for our case study on Stable Diffusion 2~\cite{rombach2022high} (\autoref{sec:appendix_vis}).

\appendix
\section{Proof for \autoref{th:theo1}}\label{sec:proof}

We start our analysis from ideal generative model and
reconstruction algorithm, which we define as deterministic
generative model and perfect reconstruction algorithm:

\begin{definition}[Deterministic Generative Model]\label{def:deterministic}
    Given a generative model \(\mathcal{M}: \mathcal{I}
    \mapsto \mathcal{X}_{\mathcal{M}}\), it is deterministic
    if it always produce the same output \(\bm x \in
    \mathcal{X}_{\mathcal{M}}\) given the same input \(\bm i
    \in \mathcal{I}\).
\end{definition}

\begin{definition}[Perfect Reverse-engineering Algorithm]\label{def:perfect_recon}
    Given a reverse-engineering algorithm \(\mathcal{A}\), if it is guaranteed
    that the returned reconstruction loss \(l\) is the global minima, then we say
    \(\mathcal{A}\) is a perfect reverse-engineering algorithm.
\end{definition}

\begin{proof}
Assume the given output sample \(\bm x\) is generated by
input \(\bm i\). Since the given model \(\mathcal{M}\) is
deterministic, we have:
\begin{equation}
    \bm x = \mathcal{M}(\bm i)
\end{equation}
In
this case, the distance between the \(\bm x\) and
\(\mathcal{M}(\bm i)\) is 0, i.e.,
\(\mathcal{L}\left(\mathcal{M}(\bm i), \bm x\right) = 0\).
Based on \autoref{def:reconstruction}, as \(\mathcal{A}\) is
a perfect reverse-engineering algorithm, it can find the input
that can produce the minimal reconstruction loss. Therefore,
we have:
\begin{equation}
    \forall \bm x \in \mathcal{X}_\mathcal{M},\mathcal{A}(\mathcal{M}, \bm x) = 0
\end{equation}

Similarly, for sample
\(\bm x^{\prime} \notin \mathcal{X}_\mathcal{M}\). There
does not exist an input \(i^{\prime}\) that can produce
\(\bm x^{\prime}\), meaning that there does not exist an
input \(\bm i^{\prime}\) that have
\(\mathcal{L}\left(\mathcal{M}(\bm i^{\prime}), \bm
x^{\prime}\right) = 0\). Thus, we have:

\begin{equation}
    \forall \bm
    x^{\prime} \notin \mathcal{X}_\mathcal{M},
    \mathcal{A}(\mathcal{M}, \bm x^{\prime}) > 0
\end{equation}

Finally, we
have for any \(\bm x \in \mathcal{X}_{\mathcal{M}}\) and
\(\bm x^{\prime} \notin \mathcal{X}_{\mathcal{M}}\) we have
\(\mathcal{A}(\mathcal{M}, \bm x^{\prime}) >
\mathcal{A}(\mathcal{M}, \bm x)\).

\end{proof}

\section{Computing critical value of the \(t\) distribution. }\label{sec:appendix_critical}

In \autoref{sec:hypo}, we use the critical value of the \(t\) distribution to obtain the results of the hypothesis testing.
In this section, we discuss the detailed process for calculating the 
critical value.
For the t-distribution, we have the probability density function:

\begin{equation}
\label{eq:pdf}
f(t)=\frac{\Gamma\left(\frac{\nu+1}{2}\right)}{\sqrt{\nu \pi} \Gamma\left(\frac{\nu}{2}\right)}\left(1+\frac{t^2}{\nu}\right)^{-(\nu+1) / 2}
\end{equation}

In \autoref{eq:pdf},
\(\nu\)  is the number of degrees of freedom and 
\(\Gamma\)  is the gamma function. Based on \autoref{eq:pdf}, we have the cumulative distribution function:

\begin{equation}
\mathbb{P}(t<t^{\prime})=\int_{-\infty}^{t^{\prime}} f(u) d u=1-\frac{1}{2} \beta\left(\frac{\nu}{{t^{\prime}}^2 + \nu},\frac{\nu}{2}, \frac{1}{2}\right)
\end{equation}

where \(\beta\) denotes the incomplete beta function. 
Therefore, given a confidence level \(\alpha\) and the number of degrees of freedom \(\nu\), we have can use \autoref{eq:tav} to obtain the value of the critical value \(t_{\alpha, \nu}\).

\begin{equation}
\label{eq:tav}
\mathbb{P}(t<t_{\alpha, \nu}) = 1 - \alpha
\end{equation}

\section{More Results for Distinguishing Belonging Images and Images Generated by Other Models. }\label{sec:appendix_trainingdata}
In this section, we provide more results for distinguishing belonging images of the model and the images generated by other models. 
Given a model \(\mathcal{M}_1\), we focus on the other model \(\mathcal{M}_2\)
which shares the same architecture as \(\mathcal{M}_1\)
  and has training data that overlaps with \(\mathcal{M}_2\)
 's training data.
 \begin{wraptable}{r}{0.65\linewidth}
\centering
\scriptsize
\vspace{-0.1cm}
\caption{Results when the examined model and the other model has the same architecture, and their training data has overlaps.}\label{tab:overlap}
\begin{tabular}{@{}cc@{}}
\toprule
Overlap Fraction & Acc    \\ \midrule
50\%             & 98.5\% \\
60\%             & 98.0\% \\
70\%             & 98.5\% \\
80\%             & 96.0\% \\
90\%             & 96.5\% \\ \bottomrule
\end{tabular}
\vspace{-0.1cm}
\end{wraptable}
 The model architecture used here is DCGAN~\cite{radford2015unsupervised}. We trained \(\mathcal{M}_1\) on the full CIFAR-10~\cite{krizhevsky2009learning} dataset, while \(\mathcal{M}_2\) is trained on the randomly sampled subsets of the CIFAR-10 dataset.
The results are presented in \autoref{tab:overlap}, where the first column indicates the proportion of overlap between the training data of \(\mathcal{M}_1\) and \(\mathcal{M}_2\). The second column of Table \ref{tab:overlap} displays the accuracy of \sys in detecting the differences. Notably, even when 90\% of \(\mathcal{M}_2\)'s training data overlaps with that of  \(\mathcal{M}_1\), the detection accuracy remains above 95\%. These results demonstrate that our method is still effective when 
the training data of the examined model and that of the other model are similar.

\section{Adaptive Attack}\label{sec:appendix_editing}

In this section, we evaluate the robustness of \sys against the adaptive attack where the malicious user is aware of it and try to bypass the inspection of \sys. For example, when the malicious user use a specific model to generate an image, he can make a slight modification on the image to bypass the inspection of the origin attribution algorithm.
\begin{wraptable}{r}{0.49\linewidth}
\centering
\scriptsize
\setlength\tabcolsep{3pt}
\caption{Results under adaptive attack.}\label{tab:adaptive}
\begin{tabular}{@{}ccccc@{}}
\toprule
TP & FP & FN & TN & Acc    \\ \midrule
96 & 15 & 4  & 85 & 90.5\% \\ \bottomrule
\end{tabular}
\end{wraptable}
We consider the image editing such as adding an image filter as the adaptive attack because it can preserve most of the information in the original image while changing the image. To investigate if \sys is robust to the image-editing-based adaptive attack, we use the \_1977 instagram filter\footnote{https://github.com/akiomik/pilgram} to conduct results. The model used here is the DCGAN~\cite{radford2015unsupervised} model trained on the CIFAR-10~\cite{krizhevsky2009learning} dataset.
The results are shown in \autoref{tab:adaptive}. We can see that the detection accuracy of \sys is still above 90\% even under the image-editing-based adaptive attack, demonstrating the robustness of our method.

\vspace{-0.1cm}
\section{Efficiency}\label{sec:appendix_efficiency}
\vspace{-0.2cm}
In this section, we discuss the efficiency of \sys. To study the efficiency, we measure the runtime of our method on StyleGAN2-ADA~\cite{karras2020training} model trained on the CIFAR-10~\cite{krizhevsky2009learning} dataset, as well as the Consistency Model~\cite{song2023consistency} trained on the ImageNet~\cite{russakovsky2015imagenet} dataset. The average running time for StyleGAN2-ADA and the Consistency Model are 55.16s and 152.83s, respectively. Our method can be accelerated by using 
mixed precision training~\citep{micikevicius2017mixed}.
Further approach for speeding up the input reverse-engineering process will be our future work.

\vspace{-0.1cm}
\section{More 
Visualizations}\label{sec:appendix_vis}
\begin{figure*}[t]
    \captionsetup[subfigure]{labelformat=empty}
    \centering
    \subfloat[(a) The reconstruction loss and the calibrated reconstruction loss are 0.0006 and 0.0167, respectively.]%
    {
    \subfloat[Belonging]{\includegraphics[width=0.45\columnwidth]{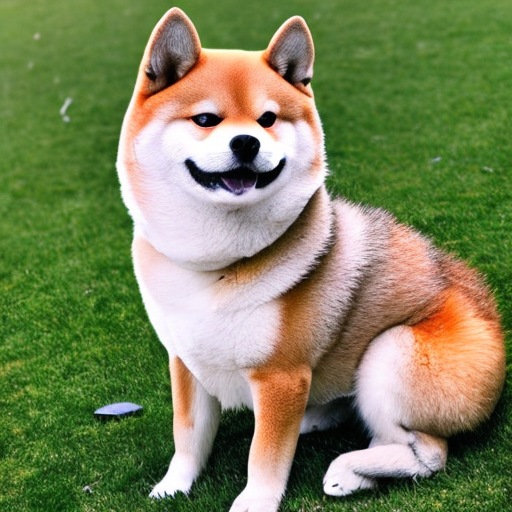}}
    \hspace{0.5cm}
    \subfloat[Reverse-engineered]{\includegraphics[width=0.45\columnwidth]{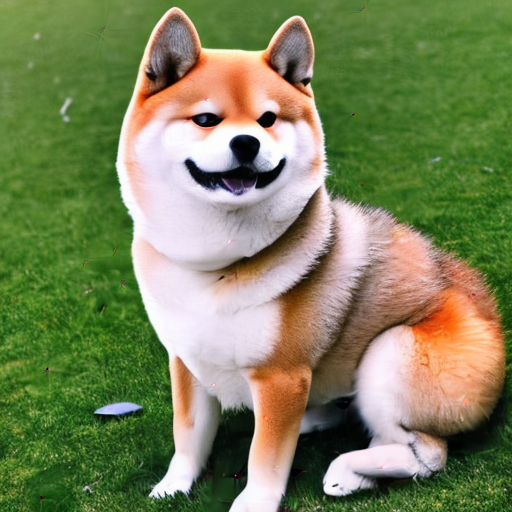}}
    }
    \hfill
    \subfloat[(b) The reconstruction loss and the calibrated reconstruction loss are 0.0008 and 0.0286, respectively.]%
    {
    \subfloat[Belonging]{\includegraphics[width=0.45\columnwidth]{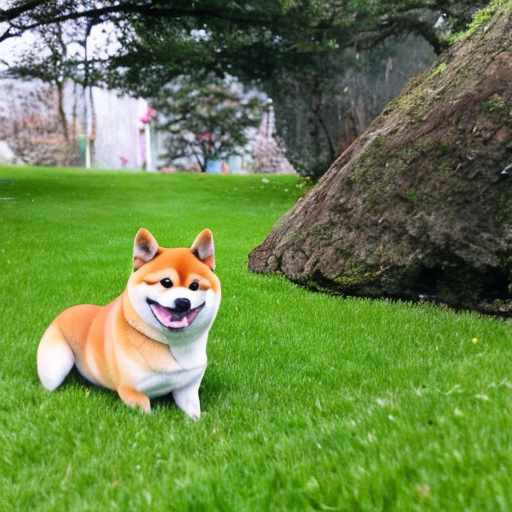}}
    \hspace{0.5cm}
    \subfloat[Reverse-engineered]{\includegraphics[width=0.45\columnwidth]{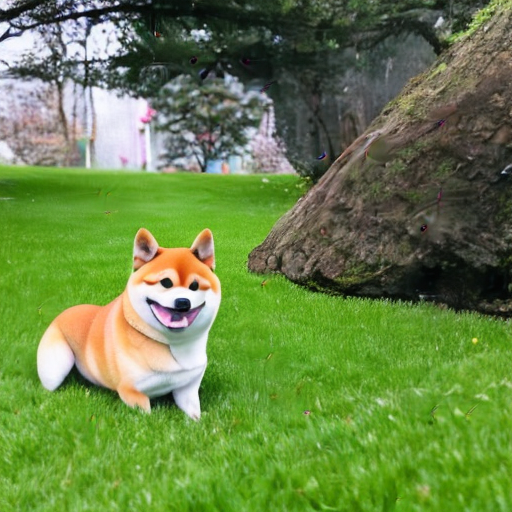}}
    }
    \hfill
    \subfloat[(c) The reconstruction loss and the calibrated reconstruction loss are 0.0005 and 0.0161, respectively.]%
    {
    \subfloat[Belonging]{\includegraphics[width=0.45\columnwidth]{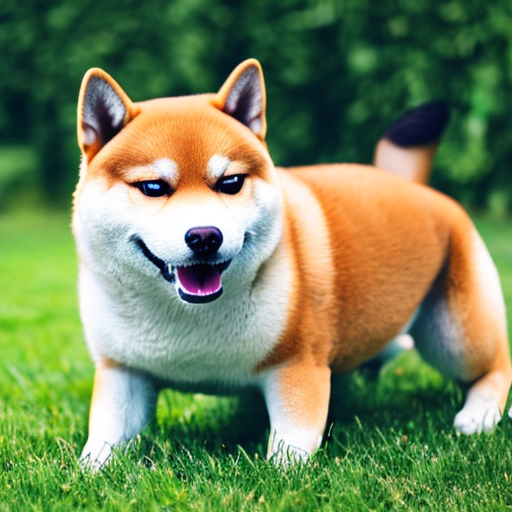}}
    \hspace{0.5cm}
    \subfloat[Reverse-engineered]{\includegraphics[width=0.45\columnwidth]{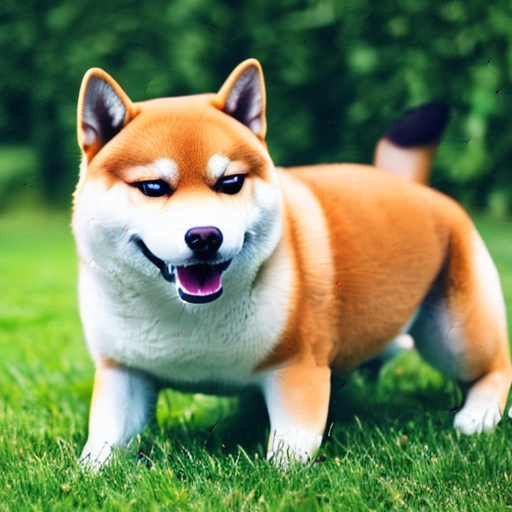}}
    }
    \vspace{0.3cm}
    \caption{More visualization of the belonging images for Stable Diffusion 2.0~\cite{rombach2022high}, and their corresponding reverse-engineered images. The distance metric used here is MSE.}\label{fig:appendix_belonging}
\end{figure*}

\begin{figure*}[t]
    \captionsetup[subfigure]{labelformat=empty}
    \centering
    \subfloat[(a) The reconstruction loss and the calibrated reconstruction loss are 0.0038 and 0.1727, respectively.]%
    {
    \subfloat[Non-belonging]{\includegraphics[width=0.45\columnwidth]{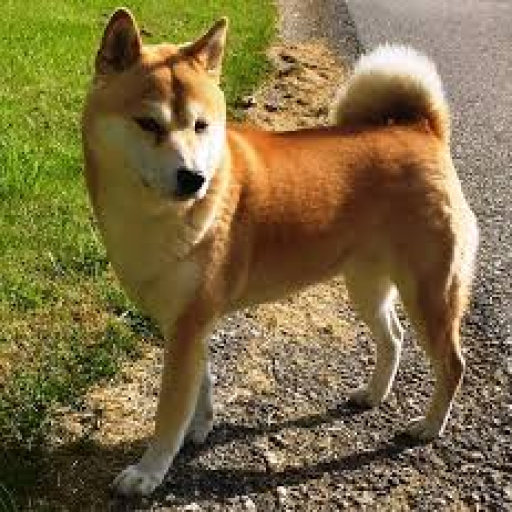}}
    \hspace{0.5cm}
    \subfloat[Reverse-engineered]{\includegraphics[width=0.45\columnwidth]{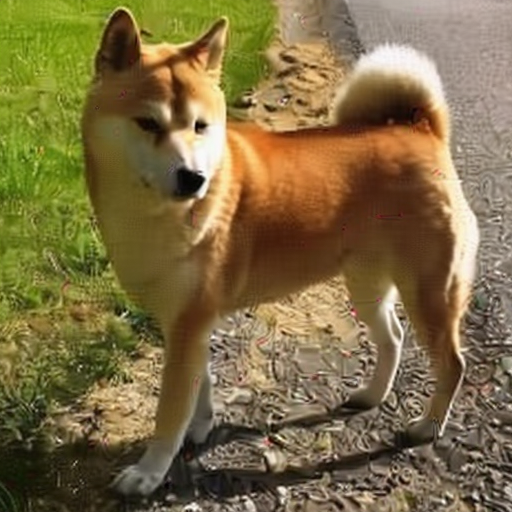}}
    }
    \hfill
    \subfloat[(b) The reconstruction loss and the calibrated reconstruction loss are 0.0019 and 0.0704, respectively.]%
    {
    \subfloat[Non-belonging]{\includegraphics[width=0.45\columnwidth]{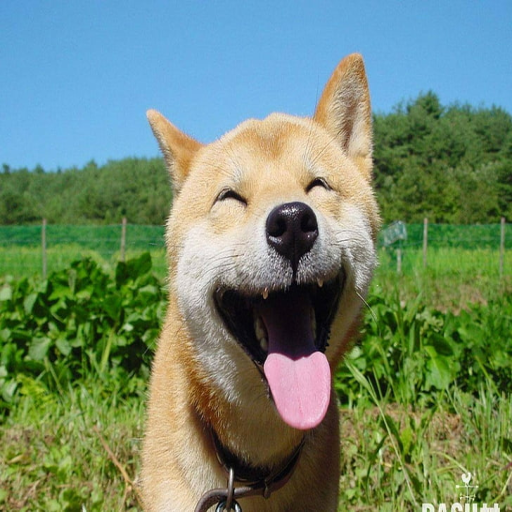}}
    \hspace{0.5cm}
    \subfloat[Reverse-engineered]{\includegraphics[width=0.45\columnwidth]{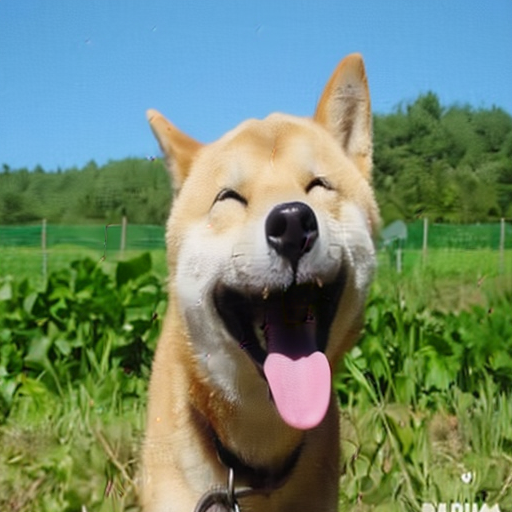}}
    }
    \hfill
    \subfloat[(c) The reconstruction loss and the calibrated reconstruction loss are 0.0028 and 0.2800, respectively.]%
    {
    \subfloat[Non-belonging]{\includegraphics[width=0.45\columnwidth]{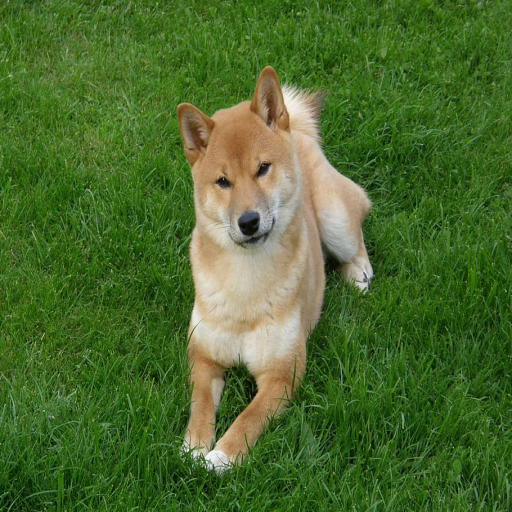}}
    \hspace{0.5cm}
    \subfloat[Reverse-engineered]{\includegraphics[width=0.45\columnwidth]{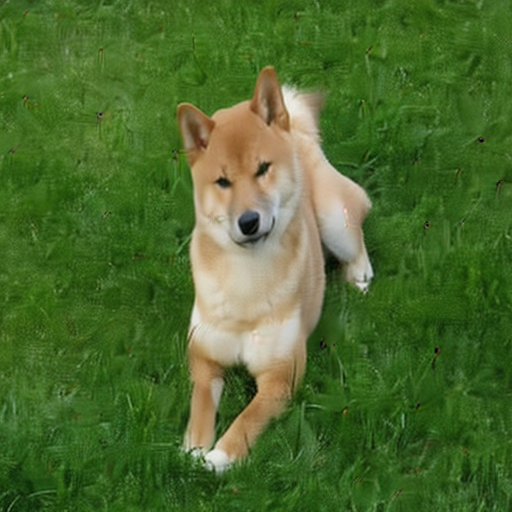}}
    }
    \vspace{0.3cm}
    \caption{More visualization of the non-belonging images for Stable Diffusion 2.0~\cite{rombach2022high}, and their corresponding reverse-engineered images. The distance metric used here is MSE.}\label{fig:appendix_nonbelonging}
\end{figure*}

\vspace{-0.2cm}
In this section, we provide more visualizations for our case study on Stable Diffusion 2.0 model~\cite{rombach2022high} (\autoref{sec:sd}).
We show more visualizations for the belonging images and their corresponding reverse-engineered images in \autoref{fig:appendix_belonging}. In \autoref{fig:appendix_nonbelonging}, we demonstrate more visualizations for the non-belonging images of the Stable Diffusion 2.0 model, and also show their reverse-engineered images. The detailed reconstruction loss and the calibrated reconstruction loss are reported in \autoref{fig:appendix_belonging} and \autoref{fig:appendix_nonbelonging}. The distance metric used here is MSE. As we can observed, the belonging images have much lower calibrated reconstruction loss than the non-belonging images, further demonstrating the effectiveness of our approach.

\end{document}